\documentclass[11pt]{article}

\usepackage[preprint]{acl}

\usepackage{times}
\usepackage{latexsym}
\usepackage[T1]{fontenc}
\usepackage[utf8]{inputenc}
\usepackage{microtype}
\usepackage{inconsolata}
\usepackage{graphicx}
\usepackage{booktabs}
\usepackage{colortbl,xcolor}
\usepackage{amsmath,amssymb}
\usepackage{multirow}

\usepackage{algorithm}
\usepackage{algpseudocode}
\usepackage{amssymb}

\newcommand{\perfgain}[1]{%
  {\scriptsize\textcolor{red}{($\uparrow$#1)}}%
}

\newcommand{\perfloss}[1]{%
  {\scriptsize\textcolor{red}{($\downarrow$#1)}}%
}

\newcommand{\effval}[2]{%
  #1 {\scriptsize\textcolor{red}{($\times$#2)}}%
}

\title{Trace, Verify, and Correct: A Training-Free Framework for Spatial Reasoning in Multimodal LLMs}

\author{
  \textbf{Yang Yang\textsuperscript{1\thanks{Equal contribution.}}},
  \textbf{Jiawei Chen\textsuperscript{1,2\footnotemark[1]}},
  \textbf{Tairan Chen\textsuperscript{3}},
  \textbf{Zhaoxia Yin\textsuperscript{1}\thanks{Corresponding author.}}
  \\
  \\
  \textsuperscript{1}Shanghai Key Laboratory of Multidimensional Information Processing,
  East China Normal University
  \\
  \textsuperscript{2}Zhongguancun Academy
  \\
  \textsuperscript{3}Stevens Institute of Technology
  \\
}

\begin{document}
\maketitle


\begin{abstract}
Although Multimodal Large Language Models (MLLMs) have made substantial progress, their spatial reasoning may still produce intermediate judgments inconsistent with the input image, allowing errors to propagate through the reasoning chain and affect the final answer. Existing methods mainly improve spatial reasoning through training or additional spatial information, without considering whether the reasoning process itself is faithful to the model input. Our study shows that unfaithful reasoning chains significantly reduce final-answer accuracy. To address this issue, we propose a modular and training-free framework for spatial reasoning verification and correction. The framework constructs a Spatial Evidence Graph (SEG), which associates atomic spatial evidence extracted from Chain-of-Thought reasoning with visual entities, spatial relations, source steps, and visual evidence. Spatial Evidence Reliability Assessment (SERA) evaluates the reliability of visual evidence based on object existence, localization, and geometric measurements. The framework then identifies the earliest spatial evidence unit contradicted by reliable visual evidence and guides the original MLLM to revise the subsequent reasoning and final answer. Across 15 model--dataset settings, our method achieves an average accuracy of 68.94\%, outperforming the compared baselines by 8.55 percentage points on average. Our code will be open-sourced.
\end{abstract}

\section{Introduction}




MLLMs have recently achieved substantial progress in visual question answering~\cite{kuang2025natural, zhong2026focus}, image understanding~\cite{jiang2026m3cotbench, meng2025mmiu}, and cross-modal reasoning~\cite{ma2026beyond, jiang2025corvid}. Nevertheless, spatial reasoning remains a major challenge. To solve complex spatial problems, MLLMs commonly employ Chain-of-Thought (CoT) reasoning to decompose the problem into a sequence of intermediate steps before deriving the final answer. However, an intermediate spatial judgment that is inconsistent with the visual input may be reused as a premise in subsequent steps, allowing local errors to propagate through the reasoning chain and affect the final answer. Reliable spatial reasoning is therefore essential for extending multimodal models from static image understanding to interaction with real-world environments~\cite{chen2026tex3d, ding2025invisible, liu2024exploring}. For example, in high-stakes domains such as autonomous driving and robotics, incorrect spatial judgments may lead to erroneous decisions, resulting in substantial economic losses or even threats to human safety.

\begin{figure}[t]
  \centering
  \includegraphics[width=\columnwidth]{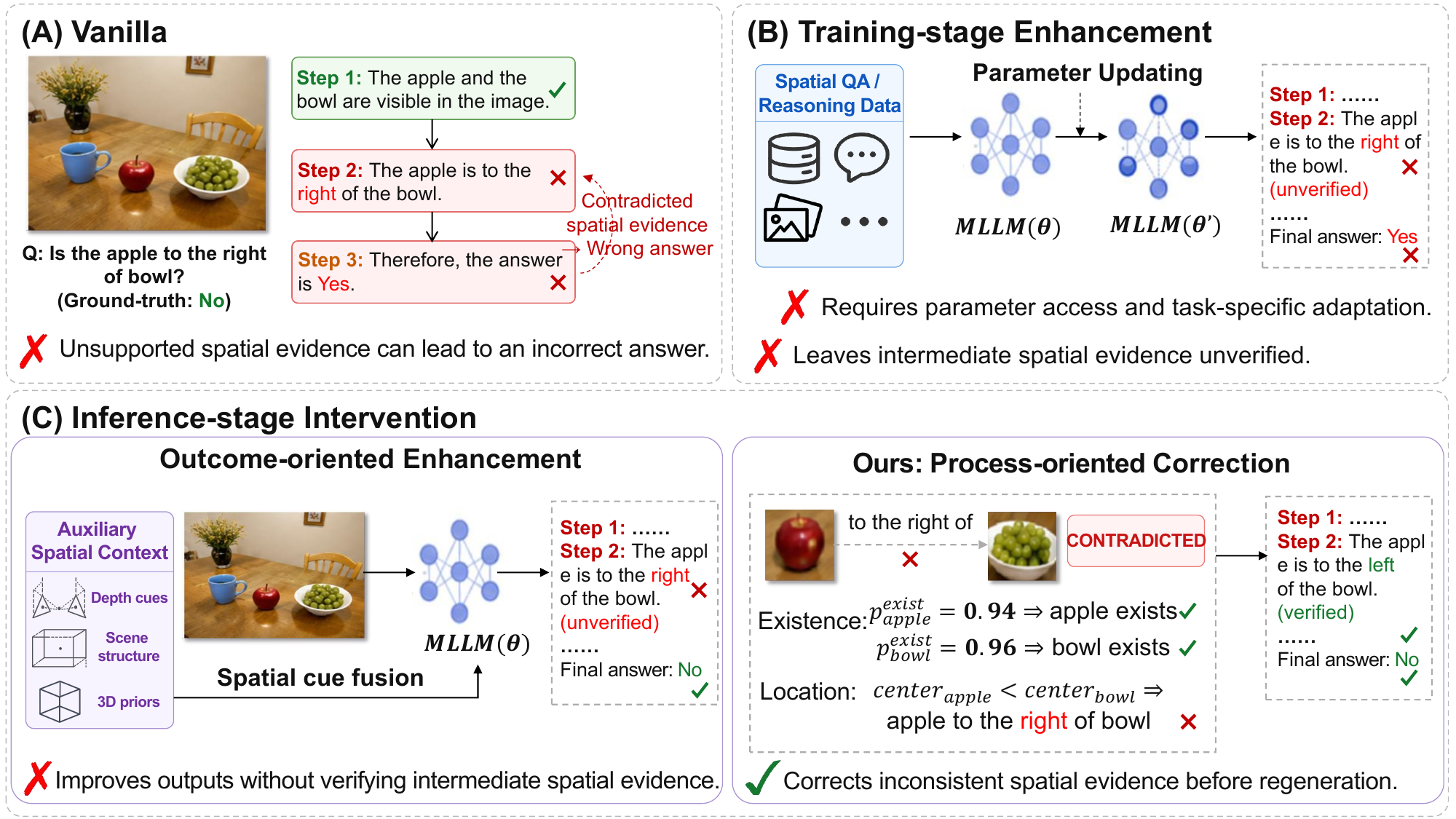}
\caption{
Motivation for perceptually faithful spatial reasoning.
(A) Vanilla reasoning propagates unsupported evidence to an incorrect answer.
(B) Training-stage methods require parameter access and task-specific adaptation.
(C) Inference-stage methods improve outputs but leave intermediate spatial evidence unverified. Our process-oriented method verifies and corrects inconsistent spatial evidence before regeneration.
}
  \label{fig:intro-motivation}
  \vspace{-2ex}
\end{figure}

As shown in Fig.~\ref{fig:intro-motivation}, prior work~\cite{gholami2025spatial,wang2025n3d,daxberger2025mm,zhang2026flatland,li2026star} enhances MLLM spatial reasoning using spatial QA or reasoning data during training, but requires parameter access and adaptation to new models or tasks. To avoid retraining, subsequent work~\cite{ma2024spatialpin,roy2025bydeway,zhu2026struct2d,li2025see,hua2026unleashing} introduces depth cues, structured scene representations, or 3D priors at inference time. 
However, these methods remain primarily outcome-oriented: they enhance the input or overall generation process without explicitly locating, verifying, or correcting intermediate spatial evidence. Consequently, intermediate reasoning errors may remain undetected and propagate to the final answer.
We term consistency between intermediate spatial evidence and visual input \emph{perceptual faithfulness}~\cite{lv2026spd,li2026faithful}. Section~\ref{sec:faithfulness_accuracy} shows that perceptually unfaithful reasoning chains have significantly lower answer accuracy. We therefore adopt a process-oriented strategy that verifies and corrects spatial evidence after initial CoT generation and regenerates the affected reasoning and final answer.

Reliably correcting such unfaithful reasoning presents two key challenges. First, spatial evidence units are difficult to extract and trace. In free-form CoT, spatial judgments may be expressed implicitly and distributed across interdependent reasoning steps, making it difficult to accurately identify each evidence unit and trace it to its source step. Second, verification evidence is not always reliable. Object existence, localization, and geometric measurements may be affected by occlusion and perception errors, and correcting the CoT based on uncertain evidence may alter otherwise valid spatial judgments and degrade answer accuracy. 

To meet these challenges, we propose a verification-and-correction framework for perceptually faithful spatial reasoning in MLLMs. SEG addresses evidence extraction and provenance tracing by representing each atomic CoT spatial judgment as an evidence unit while preserving its visual entities, spatial relations, and source step. SERA addresses unreliable verification evidence by assessing object existence, localization confidence, and geometric measurement quality. Based on these reliability-aware results, the framework identifies the earliest evidence unit contradicted by reliable visual evidence and guides the original MLLM to regenerate the affected reasoning and final answer. Correcting from the earliest reliably detected error improves perceptual faithfulness and final-answer accuracy while preserving valid reasoning. Our key contributions are:


\begin{itemize}
    \item We propose a systematic verification and correction framework for spatial reasoning in MLLMs, enabling evidence-level spatial-relation verification, error localization, and reasoning revision.
    \item We develop the SEG and SERA to support structured provenance tracing of spatial evidence units and hierarchical reliability assessment of verification evidence.
    \item Our training-free and model-agnostic framework consistently improves multiple MLLMs across spatial reasoning benchmarks, achieving an average accuracy of 68.94\% and outperforming the compared baselines by 8.55 percentage points on average.
\end{itemize}

\section{Why Reasoning-Level Spatial Correction Is Necessary}

\begin{table}[t]
\centering
\small

\setlength{\tabcolsep}{4.5pt}
\renewcommand{\arraystretch}{1.12}

\resizebox{\columnwidth}{!}{%
\begin{tabular}{llccc}
\toprule
\textbf{Dataset}
& \textbf{CoT Faithfulness}
& \textbf{Acc. (\%)}
& \textbf{OR}
& \textbf{$p$-value} \\
\midrule
\multirow{2}{*}{RealWorldQA}
& Fully Faithful & 51.10
& \multirow{2}{*}{1.99}
& \multirow{2}{*}{$<0.001$} \\
& Unfaithful & 34.41 & & \\
\bottomrule
\end{tabular}%
}
\captionsetup{skip=4pt}
\caption{
Final-answer accuracy conditioned on CoT faithfulness for
InternVL3.5-8B.
}
\label{tab:answer_reasoning_faithfulness}
\vspace{-2ex}
\end{table}


We investigate the necessity of reasoning-level spatial correction through three empirical analyses. 

\subsection{Perceptually Unfaithful Reasoning Is Associated with Lower Answer Accuracy}
\label{sec:faithfulness_accuracy}



To examine the relationship between reasoning faithfulness and answer performance, we inspect CoTs generated by InternVL3.5-8B on RealWorldQA and annotate spatial evidence units as supported, contradicted, or inconclusive. We compare faithful CoTs, in which all evidence units are supported by the image, with unfaithful CoTs containing at least one contradicted evidence unit. As shown in Tab.~\ref{tab:answer_reasoning_faithfulness}, the two groups achieve final-answer accuracies of $51.10\%$ and $34.41\%$, respectively, a difference of $16.69\%$. This difference is statistically significant (Fisher's exact test, OR $=1.99$, $p<0.001$), indicating that perceptually unfaithful spatial reasoning is significantly associated with lower final-answer accuracy. Therefore, we investigate whether improving spatial reasoning faithfulness can enhance final-answer accuracy.

\subsection{Spatial Unfaithfulness Accumulates Along the Reasoning Chain}

\begin{figure}[t]
\centering
\includegraphics[width=0.9\linewidth]{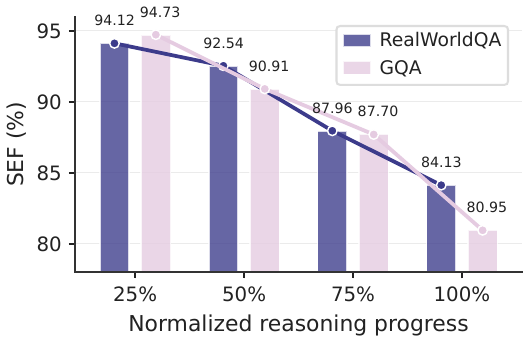}
\captionsetup{skip=4pt}
\caption{
Cumulative spatial evidence faithfulness of Vanilla on fixed 500-sample subsets of RealWorldQA and GQA using InternVL3.5-8B.
}
\label{fig:cumulative_step_faithfulness_vanilla}
\vspace{-2ex}
\end{figure}


We further normalize the position of each spatial evidence unit according to its source step and define Spatial Evidence Faithfulness at $p\%$ (SEF@$p$) as the cumulative faithfulness of evidence units originating within the first $p\%$ of the reasoning chain. As shown in Fig.~\ref{fig:cumulative_step_faithfulness_vanilla}, the SEF@$p$ of Vanilla consistently decreases as reasoning progresses on both RealWorldQA and GQA, indicating that unfaithful spatial evidence can arise at different stages and gradually accumulate. 
Therefore, improving final-answer accuracy requires locating spatial evidence units within the reasoning chain.

\subsection{Untargeted Self-Revision Is Unreliable}

\begin{figure}[t]
\centering
\includegraphics[width=\columnwidth]{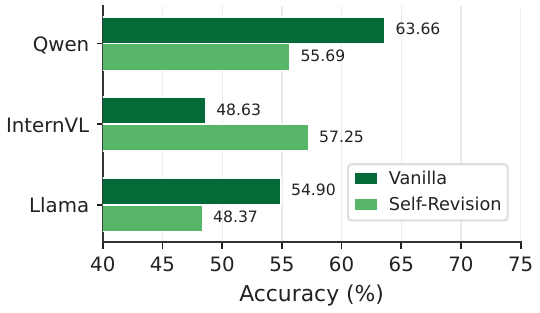}
\captionsetup{skip=4pt}
\caption{
Effect of generic Self-Revision on the final-answer accuracy of three MLLMs on RealWorldQA.
}
\label{fig:self_revision}
\vspace{-2ex}
\end{figure}

To examine whether untargeted revision of the entire reasoning chain can address these errors, we evaluate generic Self-Revision on RealWorldQA, where the same MLLM is provided with the original image, question, initial reasoning chain, and answer and is asked to reconsider and revise the complete reasoning process without explicit error localization or external visual evidence. The complete prompt is provided in Appendix~\ref{app:self_revision_prompt}. 
As shown in Fig.~\ref{fig:self_revision}, Self-Revision improves InternVL but degrades Qwen and Llama, indicating that such perceptual unfaithfulness is not merely attributable to stochastic sampling, but instead reflects systematic limitations in the visual perception of MLLMs~\cite{li2025revisor,chen2026red}. 
It is therefore necessary to introduce independent and reliable visual evidence and explicitly localize the spatial evidence units that require correction.




These analyses suggest that improving final-answer accuracy through perceptually faithful reasoning requires locating specific spatial evidence units in the CoT that conflict with the visual input and providing their locations and verification evidence for targeted regeneration.

\section{Method}

\begin{figure*}[t]
  \centering
  \includegraphics[width=0.9\linewidth]{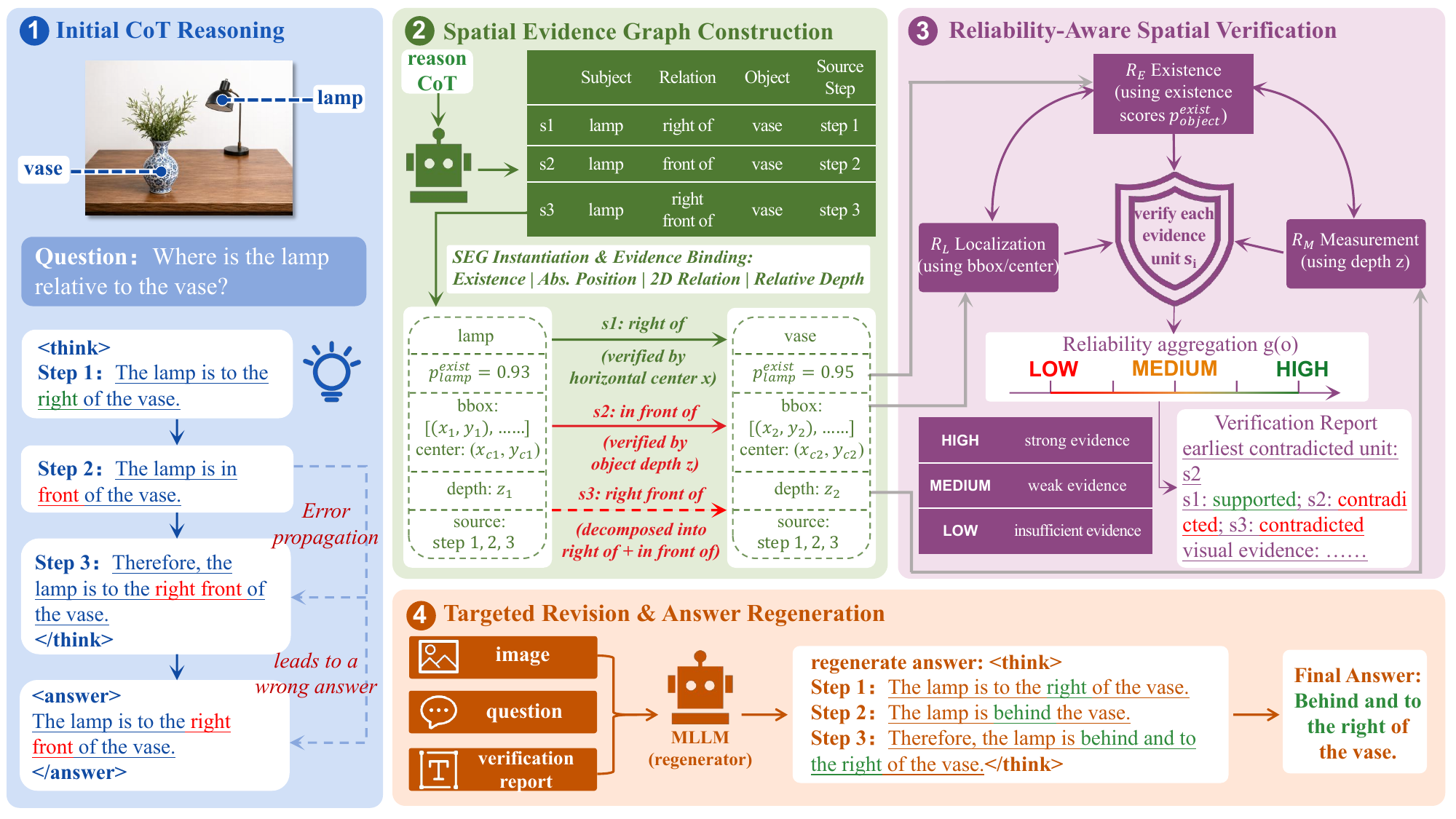}
  \captionsetup{skip=4pt}
  \caption{
    Overview of our process-oriented spatial reasoning correction framework.
    SEG structures intermediate spatial evidence, SERA assesses visual evidence reliability, and the earliest contradicted evidence unit guides targeted reasoning and answer regeneration.
    }
  \label{fig:framework}
\vspace{-2ex}
\end{figure*}



Motivated by the above analyses, we propose a training-free framework for spatial reasoning verification and correction that locates and corrects perceptually unfaithful spatial evidence units, thereby improving perceptual faithfulness and final-answer accuracy. As illustrated in Fig.~\ref{fig:framework}, given an image $I$ and a question $q$, an MLLM first generates an initial reasoning chain and answer. The framework then extracts spatial evidence units to construct a SEG, verifies each using visual evidence assessed for reliability, locates the earliest unit contradicted by reliable evidence, and guides the model to revise the affected subsequent reasoning and final answer.

\subsection{Spatial Evidence Graph Construction}

Spatial judgments in natural-language CoT are often distributed across different reasoning steps, making them difficult to verify independently and trace back to their origins when errors are detected. To address this issue, we explicitly convert these judgments into a Spatial Evidence Graph (SEG). The construction process consists of three steps: extracting spatial evidence units from the reasoning chain, mapping them into a unified graph structure, and associating them with the visual evidence required for subsequent verification.


Given an initial reasoning chain $C=\{c_1,c_2,\ldots,c_n\}$, we prompt an MLLM to extract spatial evidence units from each reasoning step across four types: object existence, absolute image position, 2D relative spatial relations, and relative depth. Each unit retains its source step, enabling unsupported evidence to be localized in the original CoT. We organize these units into a SEG $\mathcal{G}=(\mathcal{V},\mathcal{E},\mathcal{A})$, where $\mathcal{V}$ contains visual entity nodes and an image-reference node, $\mathcal{E}$ denotes directed spatial-relation edges, and $\mathcal{A}$ stores evidence types, source steps, and visual evidence. Object existence is represented as an entity-node attribute, while absolute positions and inter-object relations are represented as directed edges. Each unit is formally represented as
\vspace{-0.3em}
\[
s_m=
\left\langle
o_i,r_m,o_j,t_m,k_m
\right\rangle,
\]

where $o_i$ and $o_j$ denote the entities or reference node involved in the evidence unit, $r_m$ denotes the encoded spatial relation, $t_m$ denotes the evidence type, and $k_m$ records the source reasoning step.


Finally, we associate visual evidence by evidence type. For each entity, Grounding DINO~\cite{liu2024grounding} provides candidate boxes and confidence scores, from which we compute the existence score $p_{\mathrm{yes}}(o)$ following FaithAct. Bounding-box coordinates support absolute-position and 2D relation units. For depth relations, SAM2~\cite{ravi2025sam} provides object masks, and a monocular depth estimator~\cite{yang2024depth} computes depth statistics within them. The resulting SEG links its nodes and edges to the evidence required for reliability assessment and verification. Extraction prompts and details are provided in Appendix~\ref{app:evidence_extraction_prompt}.

\subsection{Reliability-Aware Spatial Verification}


The SEG specifies the spatial evidence units to be verified and associates them with the corresponding visual entities and evidence. However, in challenging cases involving small objects, occlusion, cluttered backgrounds, or ambiguous boundaries, the associated visual observations may exhibit substantial uncertainty. For example, object detection may produce low-confidence or incomplete bounding boxes, while depth estimation may be affected by segmentation boundary errors and depth variations within the object region. Treating all visual evidence as equally reliable in such cases may lead to overconfident verification decisions and introduce new errors into subsequent reasoning revision. We therefore introduce Spatial Evidence Reliability Assessment (SERA), which evaluates whether the relevant visual evidence is sufficiently reliable to support a decisive judgment before verifying each spatial evidence unit. This process consists of entity-level evidence reliability assessment and evidence-unit-level spatial verification.


For each visual entity $o$, SERA evaluates three complementary aspects of its evidence: existence certainty, localization clarity, and geometric-measurement stability, denoted by $R_E(o)$, $R_L(o)$, and $R_M(o)$, respectively. Let $p_{\mathrm{yes}}(o)\in[0,1]$ denote the image-level probability that entity $o$ is present, obtained during SEG construction. We define existence reliability as
\vspace{-0.3em}
\[
R_E(o)=2\left|p_{\mathrm{yes}}(o)-0.5\right|.
\]

Localization reliability measures whether the entity is localized unambiguously. Let $s_1$ and $s_2$ denote the top two raw grounding scores produced by Grounding DINO for entity $o$. We define
\vspace{-0.3em}
\[
R_L(o)=
\min\left(
\sigma(s_1),
\sigma(s_1-s_2)
\right),
\]

where the first term measures the confidence of the top-ranked candidate, while the second measures its score margin over the second-ranked candidate. A small margin indicates that multiple candidate locations have similar scores and that the localization may therefore be ambiguous. Here, $\sigma(\cdot)$ denotes the sigmoid function.

Geometric-measurement reliability evaluates the stability of the visual measurements used for evidence unit verification. For evidence units that do not involve relative depth, no additional geometric measurement is required, and we therefore set $R_M(o)=1$. For relative-depth evidence units, measurement reliability depends on both mask-area sufficiency and depth consistency:
\begin{table*}[t]
\centering
\renewcommand{\arraystretch}{1.08}

\resizebox{0.9\textwidth}{!}{%
\begin{tabular}{lcccccc}
\toprule
\textbf{Model + Method}
& \textbf{LLaVA-Bench (\%)}
& \textbf{RealWorldQA (\%)}
& \textbf{POPE (\%)}
& \textbf{GQA (\%)}
& \textbf{MMHal (\%)}
& \textbf{Average (\%)} \\
\midrule

\textbf{Qwen} + Vanilla
& 63.33
& 63.66
& 74.90
& 56.21
& 63.54
& 64.33 \\

\hspace{1.6em}+ SpatialPIN
& 58.33
& 64.44
& \underline{85.67}
& 48.33
& 63.54
& 64.06 \\

\hspace{1.6em}+ ByDeWay
& \underline{65.00}
& \underline{65.36}
& 82.90
& 56.49
& \underline{66.67}
& \underline{67.28} \\

\hspace{1.6em}+ SoM
& 60.00
& 64.18
& 85.43
& \textbf{60.10}
& 63.54
& 66.65 \\

\hspace{1.6em}+ GoM
& 56.67
& 53.20
& 85.43
& 33.88
& 54.17
& 56.67 \\

\hspace{1.6em}+ FaithAct
& 53.33
& 46.54
& 79.67
& 55.13
& 64.58
& 59.85 \\

\hspace{1.6em}+ \textbf{Ours}
& \textbf{71.67}
& \textbf{66.93}
& \textbf{86.33}
& \underline{57.20}
& \textbf{70.83}
& \textbf{70.59} \\

\midrule

\textbf{InternVL} + Vanilla
& 53.33
& 48.63
& 75.60
& 43.62
& 59.38
& 56.11 \\

\hspace{1.6em}+ SpatialPIN
& 45.00
& 60.52
& 80.83
& 51.18
& 55.21
& 58.55 \\

\hspace{1.6em}+ ByDeWay
& \underline{56.67}
& 44.18
& 85.27
& 55.22
& \underline{60.42}
& \underline{60.35} \\

\hspace{1.6em}+ SoM
& 48.33
& 56.73
& 79.30
& \underline{59.92}
& 47.92
& 58.44 \\

\hspace{1.6em}+ GoM
& 46.67
& 56.73
& \underline{86.23}
& 47.94
& 42.71
& 56.06 \\

\hspace{1.6em}+ FaithAct
& 41.67
& \underline{60.65}
& 65.63
& 49.80
& 47.92
& 53.13 \\

\hspace{1.6em}+ \textbf{Ours}
& \textbf{60.00}
& \textbf{66.14}
& \textbf{88.57}
& \textbf{60.35}
& \textbf{67.71}
& \textbf{68.55} \\

\midrule

\textbf{Llama} + Vanilla
& \underline{58.33}
& 54.90
& 81.30
& 69.06
& 54.17
& \underline{63.55} \\

\hspace{1.6em}+ SpatialPIN
& 48.33
& 45.88
& 84.73
& 66.62
& \underline{58.33}
& 60.78 \\

\hspace{1.6em}+ ByDeWay
& 51.67
& 56.60
& 86.27
& 60.26
& 55.21
& 62.00 \\

\hspace{1.6em}+ SoM
& 41.67
& 52.81
& 82.80
& 60.02
& \textbf{63.54}
& 60.17 \\

\hspace{1.6em}+ GoM
& 41.67
& 43.01
& \textbf{86.67}
& 63.00
& 47.92
& 56.45 \\

\hspace{1.6em}+ FaithAct
& 40.00
& \underline{58.43}
& \underline{86.40}
& \textbf{75.08}
& 53.13
& 62.61 \\

\hspace{1.6em}+ \textbf{Ours}
& \textbf{65.00}
& \textbf{61.18}
& 85.30
& \underline{71.72}
& 55.21
& \textbf{67.68} \\

\bottomrule
\end{tabular}
    }
    \captionsetup{skip=4pt}
\caption{
Performance comparison across five multimodal reasoning benchmarks using Qwen3-VL-8B-Instruct,
InternVL3.5-8B, and Llama-3.2-11B-Vision-Instruct.
Bold and underlined values denote the best and second-best results, respectively, within each model--dataset setting.
}
\label{tab:main_results}
\vspace{-2ex}
\end{table*}
\[
R_M(o)=
\begin{cases}
1,\ \text{for non-depth evidence units},\\[1mm]
a_{\mathrm{mask}}(o)\,d_{\mathrm{stable}}(o).
\end{cases}
\]
\vspace{-1ex}
\[
a_{\mathrm{mask}}(o)=
\sigma\left(
\log_{10}\left(\max(|M_o|,1)\right)-\theta_m
\right),
\]
\vspace{-1ex}
\[
d_{\mathrm{stable}}(o)=
\sigma\left(
\theta_d-\operatorname{Std}(D_o)
\right).
\]

Here, $M_o$ denotes the segmentation mask of entity $o$, and $D_o$ denotes the depth values within the corresponding masked region. A larger valid mask region and lower within-region depth variance indicate a more stable depth measurement.

The three reliability components are aggregated into an entity-level evidence reliability score using a weighted geometric mean:
\[
g(o)=
R_E(o)^{\alpha}
R_L(o)^{\beta}
R_M(o)^{\gamma},
\quad
\alpha+\beta+\gamma=1.
\]
\vspace{-1ex}

This aggregation substantially reduces the overall reliability when any critical evidence component has a low value, while $\alpha$, $\beta$, and $\gamma$ control the relative contributions of the three components. Based on $g(o)$, the evidence associated with each entity is divided into three reliability tiers:
\vspace{-0.3ex}
\[
\operatorname{tier}(o)=
\begin{cases}
\mathrm{HIGH},
& g(o)\geq\tau_h,\\
\mathrm{MEDIUM},
& \tau_l\leq g(o)<\tau_h,\\
\mathrm{LOW},
& g(o)<\tau_l.
\end{cases}
\]
\vspace{-0.3ex}



We set $\tau_h=0.75$ and $\tau_l=0.45$. A positive object-existence unit is marked as \textsc{Supported} if $R_E(o)\geq\tau_h$ and $p_{\mathrm{yes}}(o)>0.5$, and \textsc{Contradicted} if $R_E(o)\geq\tau_h$ and $p_{\mathrm{yes}}(o)<0.5$; labels are reversed for negative units. Any existence unit with $R_E(o)<\tau_h$ is marked as \textsc{Inconclusive}.

After verification, we sort the results by source step and locate the earliest spatial evidence unit contradicted by reliable visual evidence. Its source step, result, and visual evidence form a verification report, which is provided with the original image and question to the same MLLM to revise the judgment and regenerate the affected reasoning and final answer. If no reliable contradiction is found, the original reasoning and answer are retained. Detailed verification rules and the regeneration prompt are provided in Appendix~\ref{app:reasoning_regeneration_prompt}.





\section{Experiments}

\subsection{Experimental Setup}

\paragraph{Datasets and models.}

We evaluate our method on five benchmarks: 60 samples from LLaVA-Bench-in-the-Wild~\cite{liu2023visual}, 765 samples from RealWorldQA~\cite{ai2024grok}, 3,000 samples from the POPE split~\cite{li2023evaluating}, 12,578 samples from the GQA balanced test-dev~\cite{hudson2019gqa}, and 96 samples from MMHal-Bench~\cite{sun2024aligning}. These benchmarks cover open-ended visual question answering, real-world scene understanding, object-existence hallucination, and spatial-relation reasoning. 
Evaluation protocols for different datasets are provided in Appendix~\ref{app:answer_evaluation_prompts}.
We use Qwen3-VL-8B-Instruct~\cite{bai2025qwen3}, InternVL3.5-8B~\cite{wang2025internvl3}, and Llama-3.2-11B-Vision-Instruct~\cite{grattafiori2024llama} as backbone models.


\paragraph{Baselines and metrics.}


We compare our method with Vanilla, SpatialPIN~\cite{ma2024spatialpin}, ByDeWay~\cite{roy2025bydeway}, SoM~\cite{yang2023set}, GoM~\cite{frisoni2026graph}, and FaithAct~\cite{li2026faithful}. All methods use identical samples and evaluation procedures, with final-answer accuracy as the primary metric. We also use Spatial Evidence Faithfulness (SEF) to assess intermediate reasoning. For sample $i$, let $S_i$ and $C_i$ denote the numbers of \textsc{Supported} and \textsc{Contradicted} units, respectively, and define $\mathrm{SEF}_i=S_i/(S_i+C_i)$. 



\subsection{Main Results}


To evaluate the overall effectiveness of our framework and its applicability across models and datasets, we compare it with Vanilla reasoning and several inference-time spatial augmentation and verification methods on three MLLMs and five benchmarks. As shown in Tab.~\ref{tab:main_results}, our method achieves the best performance in 11 of the 15 model--dataset settings. Specifically, the average accuracy increases from $64.33\%$ to $70.59\%$ for Qwen, from $56.11\%$ to $68.55\%$ for InternVL, and from $63.55\%$ to $67.68\%$ for Llama. In contrast, the other methods exhibit more variable gains across models and benchmarks. These results demonstrate that our method provides consistent improvements across different backbones and datasets, indicating that explicitly verifying and correcting intermediate spatial evidence is more effective than merely augmenting the initial spatial information.

\subsection{Reasoning Faithfulness}

\begin{table}[t]
\centering
\small
\setlength{\tabcolsep}{4.2pt}
\renewcommand{\arraystretch}{1.18}

\resizebox{\columnwidth}{!}{%
\begin{tabular}{lccc}
\toprule
\multirow{2}{*}{\textbf{Dataset}}
& \multicolumn{2}{c}{\textbf{Spatial Evidence Faithfulness (\% $\uparrow$)}}
& \multirow{2}{*}{
    \shortstack{\textbf{Final-answer}\\
    \textbf{Accuracy (\% $\uparrow$)}}} \\
\cmidrule(lr){2-3}
& \textbf{Human Eval.}
& \textbf{Model Eval.}
& \\
\midrule

RealWorldQA
& \shortstack{
    82.12 $\rightarrow$ \textbf{92.47}\\
    \textcolor{red}{\scriptsize $(+10.35)$}
}
& \shortstack{
    81.79 $\rightarrow$ \textbf{91.40}\\
    \textcolor{red}{\scriptsize $(+9.61)$}
}
& \shortstack{
    45.00 $\rightarrow$ \textbf{57.00}\\
    \textcolor{red}{\scriptsize $(+12.00)$}
} \\

\midrule

GQA
& \shortstack{
    90.94 $\rightarrow$ \textbf{96.70}\\
    \textcolor{red}{\scriptsize $(+5.76)$}
}
& \shortstack{
    90.03 $\rightarrow$ \textbf{95.60}\\
    \textcolor{red}{\scriptsize $(+5.57)$}
}
& \shortstack{
    62.00 $\rightarrow$ \textbf{69.00}\\
    \textcolor{red}{\scriptsize $(+7.00)$}
} \\

\bottomrule
\end{tabular}%
}

\vspace{2pt}
\captionsetup{skip=4pt}
\caption{
Relationship between spatial evidence faithfulness and final-answer accuracy on 200-example subsets of RealWorldQA and GQA.
Values before and after $\rightarrow$ correspond to Vanilla and our method, respectively.
}
\label{tab:faithfulness_accuracy_relation}
\vspace{-2ex}
\end{table}

\begin{figure}[t]
\centering
\includegraphics[width=0.9\linewidth]{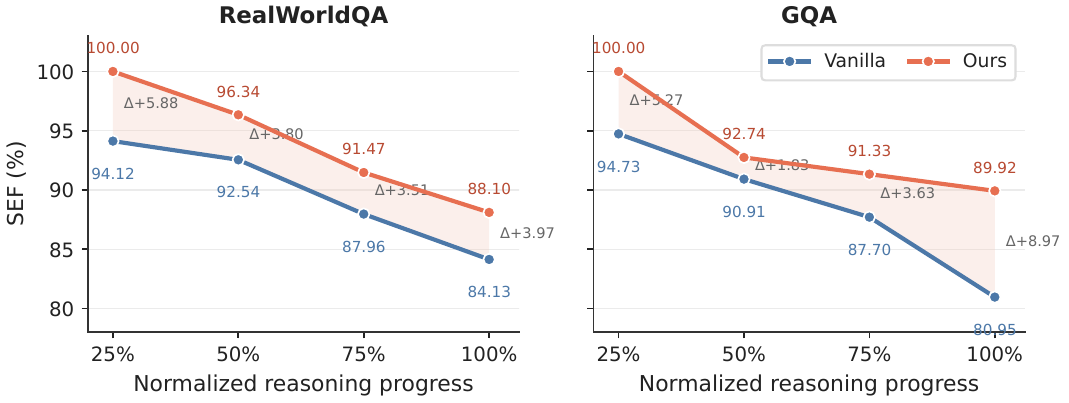}
\captionsetup{skip=4pt}
\caption{
Cumulative spatial evidence faithfulness on fixed 500-sample subsets of RealWorldQA and GQA using InternVL3.5-8B. 
}
\label{fig:cumulative_step_faithfulness}
\vspace{-2ex}
\end{figure}

To examine whether our method improves perceptual faithfulness, we evaluate SEF on fixed subsets of RealWorldQA and GQA and analyze its change over the reasoning process. As shown in Tab.\ref{tab:faithfulness_accuracy_relation}, both human and automatic evaluations show an upward trend in SEF, accompanied by improvements in final-answer accuracy. Figure~\ref{fig:cumulative_step_faithfulness} shows that Vanilla's SEF steadily declines as reasoning progresses, whereas our method remains consistently higher and improves SEF@100 by $8.97\%$ on GQA. These results demonstrate that the framework improves both the perceptual faithfulness of intermediate reasoning and final-answer accuracy.

\subsection{Ablation Studies}

\begin{table}[t]
\centering
\footnotesize
\renewcommand{\arraystretch}{1.15}
\setlength{\tabcolsep}{3.5pt}

\begin{tabular*}{\columnwidth}{
@{\extracolsep{\fill}}
lccc
@{}
}
\toprule
\textbf{Backbone}
& \textbf{Full}
& \textbf{w/o SEG} {\scriptsize($\Delta$)}
& \textbf{w/o SERA} {\scriptsize($\Delta$)} \\
\midrule

Qwen3-VL-8B
& \textbf{66.93}
& 63.92 {\scriptsize\textcolor{red}{($-3.01$)}}
& 62.88 {\scriptsize\textcolor{red}{($-4.05$)}} \\

InternVL3.5-8B
& \textbf{66.14}
& 63.01 {\scriptsize\textcolor{red}{($-3.13$)}}
& 61.96 {\scriptsize\textcolor{red}{($-4.18$)}} \\

Llama-3.2-11B
& \textbf{61.18}
& 58.56 {\scriptsize\textcolor{red}{($-2.62$)}}
& 57.65 {\scriptsize\textcolor{red}{($-3.53$)}} \\

\bottomrule
\end{tabular*}
\captionsetup{skip=4pt}
\caption{
Ablation study of SEG and SERA on RealWorldQA across multiple MLLMs. 
}
\label{tab:multi_model_ablation}
\vspace{-2ex}
\end{table}

\paragraph{Component ablation.}
To assess the contributions of the two core modules, we remove SEG and SERA separately. In w/o SEG, spatial evidence units are processed as a flat list without constructing a graph connecting entities, relations, visual evidence, and source steps. In w/o SERA, SEG and visual evidence collection are retained, but reliability assessment and gating are removed, allowing direct evidence use for verification and correction. As shown in Tab.~\ref{tab:multi_model_ablation}, removing SEG and SERA reduces average accuracy across the three backbones by $2.92$ and $3.92$ percentage points, respectively, confirming the contributions of structured evidence organization and reliability-aware verification.

\begin{figure}[t] 
\centering 
\includegraphics[width=\columnwidth]{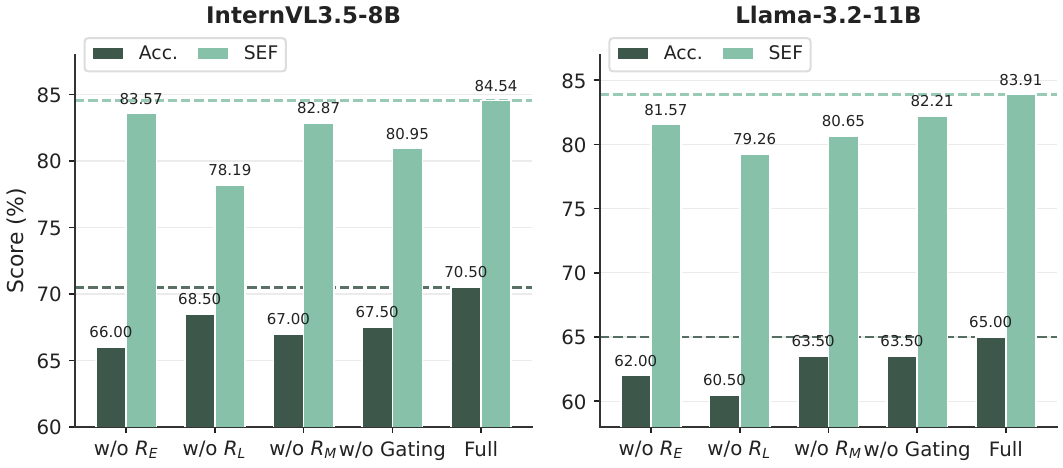} 
\captionsetup{skip=4pt}
\caption{Fine-grained ablation of SERA on a fixed 200-sample subset of RealWorldQA. 
} 
\label{fig:sera_fine_ablation} 
\vspace{-2ex}
\end{figure}

\paragraph{Fine-grained SERA ablation.}
To analyze SERA's components, we separately remove $R_E$, $R_L$, $R_M$, and reliability-aware gating. In w/o $R_E$, only $R_L$ and $R_M$ are aggregated with renormalized weights. In w/o $R_L$ or w/o $R_M$, the corresponding visual evidence is retained but excluded from reliability estimation. In w/o Gating, all components remain, but reliability tiers no longer determine whether verification is decisive. As shown in Fig.~\ref{fig:sera_fine_ablation}, each removal reduces accuracy and SEF, with $R_L$ having the largest impact, confirming their complementary roles.

\begin{table}[t]
\centering
\footnotesize
\setlength{\tabcolsep}{5pt}
\renewcommand{\arraystretch}{1.12}

\begin{tabular*}{\columnwidth}{
@{\extracolsep{\fill}}lcccc@{}
}
\toprule
\textbf{Setting}
& $\boldsymbol{\tau_l}$
& $\boldsymbol{\tau_h}$
& \textbf{SEF (\%)}
& \textbf{Acc. (\%)} \\
\midrule

Loose
& 0.35
& 0.65
& 84.04
& 51 \\

Default
& 0.45
& 0.75
& \textbf{89.36}
& \textbf{55} \\

Strict
& 0.55
& 0.85
& 86.17
& 52 \\

\bottomrule
\end{tabular*}
\caption{
Sensitivity of SERA to its reliability thresholds on a
fixed 100-sample subset of RealWorldQA using
Qwen3-VL-8B-Instruct.
}
\label{tab:sera_threshold_sensitivity}
\vspace{-2ex}
\end{table}

\paragraph{Threshold sensitivity.}

To assess SERA's threshold sensitivity, we compare loose, default, and strict settings. As shown in Tab.~\ref{tab:sera_threshold_sensitivity}, the default setting achieves the highest SEF and accuracy. Loose thresholds allow noisy evidence to trigger correction, whereas strict thresholds yield more inconclusive units. The default thresholds better balance correction coverage and evidence reliability.

\subsection{Human Validation of Intermediate Components}

To validate the automatic components, we randomly sampled 50 CoT responses each from RealWorldQA and GQA. Human annotators identify spatial evidence units and their source steps for strict exact-match evaluation of the extractor, and label the resulting 291 units as \emph{Supported}, \emph{Contradicted}, or \emph{Inconclusive} for verifier evaluation. As shown in Tab.\ref{tab:component_human_validation}, the extractor achieves F1 scores of $91.99\%$ and $91.72\%$ on the two datasets, while the verifier achieves $94.85\%$ three-way accuracy and $81.76\%$ Macro-F1, with $95.47\%$ decisive accuracy at $91.07\%$ coverage, indicating high agreement with human judgments, providing a reliable basis for subsequent correction.

\begin{table}[t]
\centering
\small
\renewcommand{\arraystretch}{1.12}

\noindent\textbf{(a) Spatial Evidence Extraction}
\par\vspace{2pt}

\setlength{\tabcolsep}{5pt}
\resizebox{\columnwidth}{!}{%
\begin{tabular}{lccc}
\toprule
\textbf{Dataset}
& \textbf{Precision (\%)}
& \textbf{Recall (\%)}
& \textbf{F1 (\%)} \\
\midrule
RealWorldQA & 89.19 & 94.96 & 91.99 \\
GQA         & 88.89 & 94.74 & 91.72 \\
\bottomrule
\end{tabular}%
}

\vspace{6pt}

\noindent\textbf{(b) Automatic Spatial Verification}
\par\vspace{2pt}

\setlength{\tabcolsep}{3pt}
\resizebox{\columnwidth}{!}{%
\begin{tabular}{lccccccc}
\toprule
\textbf{Dataset}
& \textbf{3-way Acc.}
& \textbf{F1-S}
& \textbf{F1-C}
& \textbf{F1-I}
& \textbf{Macro-F1}
& \textbf{Dec. Acc.}
& \textbf{Coverage} \\
\midrule
RealWorldQA
& 94.96
& 98.20
& 58.82
& 92.31
& 83.11
& 95.00
& 85.61 \\

GQA
& 94.74
& 97.81
& 58.82
& 76.92
& 77.85
& 95.86
& 96.05 \\
\bottomrule
\end{tabular}%
}
\caption{
Human validation of the spatial evidence extractor and verifier. Panel (a) reports extraction quality under strict exact matching of all evidence fields and source steps. Panel (b) reports verification quality. S, C, and I denote \emph{Supported}, \emph{Contradicted}, and \emph{Inconclusive}. Dec. Acc. is accuracy on human-decidable units, and Coverage is the proportion assigned a decisive label.
}
\label{tab:component_human_validation}
\vspace{-2ex}
\end{table}

\subsection{Efficiency and Scope Analysis}

\begin{table}[t]
\centering
\footnotesize

\renewcommand{\arraystretch}{1.16}
\setlength{\tabcolsep}{3.5pt}

\begin{tabular*}{\columnwidth}{
@{\extracolsep{\fill}}
lccc
@{}
}
\toprule

& \multicolumn{1}{c}{\textbf{Performance}}
& \multicolumn{2}{c}{\textbf{Inference Efficiency}} \\

\cmidrule(lr){2-2}
\cmidrule(lr){3-4}

\textbf{Method}
& \shortstack{\textbf{Accuracy}\\\textbf{(\%, $\uparrow$)}}
& \shortstack{\textbf{E2E Time}\\\textbf{(s)}}
& \shortstack{\textbf{Time/Tok.}\\\textbf{(ms)}}
\\

\midrule

Vanilla
& 53.00
& 0.8943
& 1.4425
\\

SpatialPIN
& 52.00 \perfloss{1.00}
& \effval{1.3151}{1.47}
& \effval{1.0950}{0.76}
\\

ByDeWay
& 58.00 \perfgain{5.00}
& \effval{1.3155}{1.47}
& \effval{1.0262}{0.71}
\\

SoM
& 59.00 \perfgain{6.00}
& \effval{0.9626}{1.08}
& \effval{1.4232}{0.99}
\\

GoM
& 50.00 \perfloss{3.00}
& \effval{0.9384}{1.05}
& \effval{1.5132}{1.05}
\\

FaithAct
& 58.00 \perfgain{5.00}
& \effval{1.2566}{1.41}
& \effval{1.0425}{0.72}
\\

\midrule

\textbf{Ours}
& \textbf{61.00} \perfgain{8.00}
& \effval{2.5095}{2.81}
& \effval{3.3592}{2.33}
\\

\bottomrule
\end{tabular*}
\caption{
Performance and efficiency on a 100-sample GQA subset using InternVL3.5-8B.
E2E Time denotes average end-to-end inference latency, while Time/Tok. denotes the latency normalized by output length.
}
\label{tab:gqa_efficiency}
\vspace{-2ex}
\end{table}



To assess inference cost and cross-task generality, we evaluate efficiency on a fixed 100-sample GQA subset using InternVL3.5-8B and transfer to ScienceQA and MathVista across three backbones. All latency measurements use identical hardware and software settings on eight NVIDIA GeForce RTX 4090 GPUs.

\paragraph{Inference efficiency.}
As shown in Tab.~\ref{tab:gqa_efficiency}, although our method incurs higher overall latency, its absolute runtime remains on the order of seconds while delivering the largest accuracy gain of $8.00$ percentage points. Moreover, it does not modify the underlying generation function or decoding procedure of the MLLM. Thus, our method achieves a substantial performance improvement with manageable test-time overhead while preserving the original generation mechanism.

\begin{figure}[t]
\centering
\includegraphics[width=0.9\linewidth]{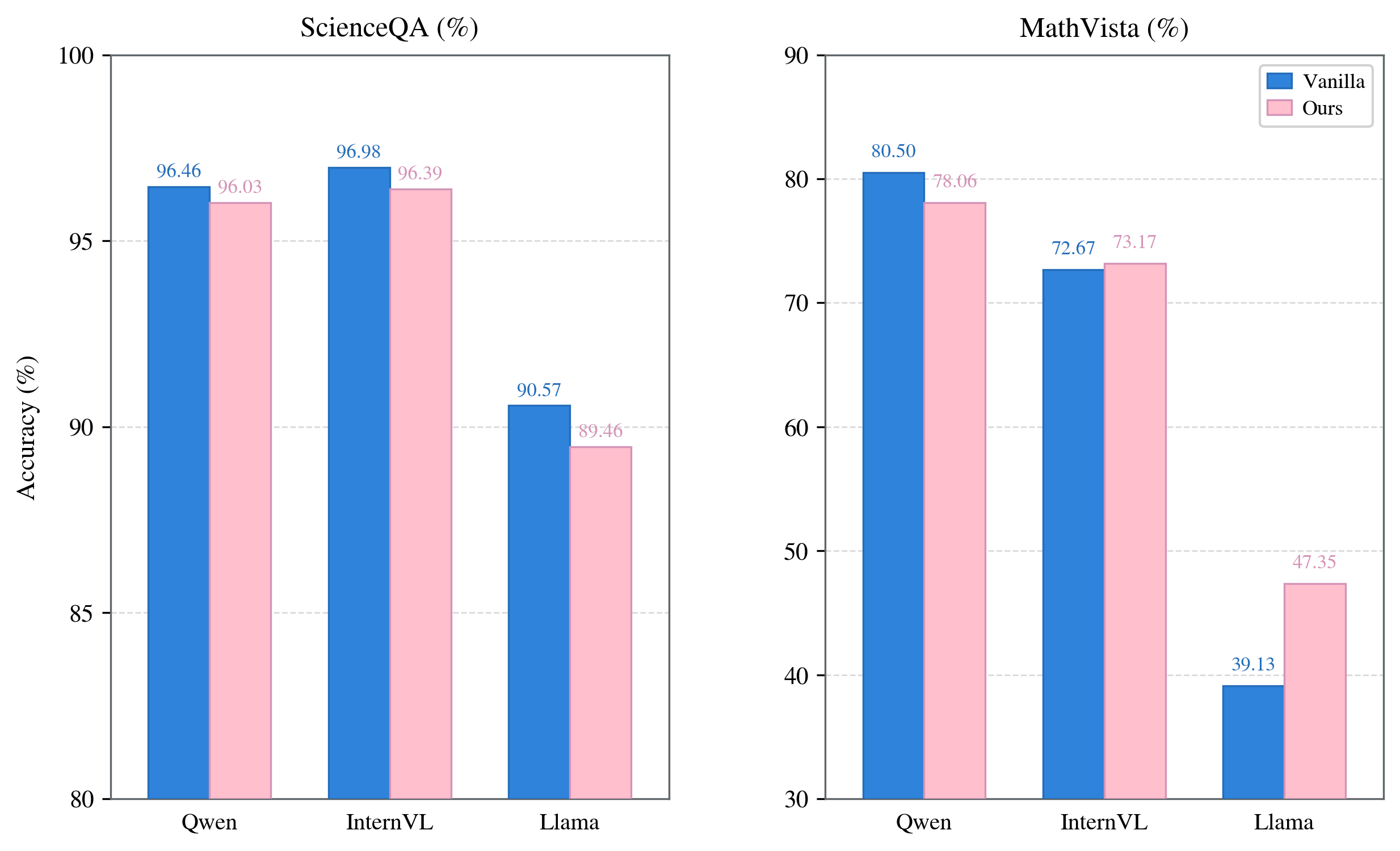}
\captionsetup{skip=4pt}
\caption{
Transfer analysis on ScienceQA and MathVista. Results are reported as final-answer accuracy.
}
\label{fig:non_spatial_transfer}
\vspace{-2ex}
\end{figure}

\paragraph{Transfer beyond spatial reasoning.}
As shown in Fig.~\ref{fig:non_spatial_transfer}, the maximum performance decrease on ScienceQA is only $1.11$ percentage points. On MathVista, Qwen decreases by $2.44$ points, whereas InternVL and Llama improve by $0.50$ and $8.22$ points, respectively. Overall, our method largely preserves the models' performance on broader multimodal tasks and provides additional gains in some settings.



\subsection{Complementarity with Inference-Time Spatial Augmentation}
\label{sec:augmentation_complementarity}


To evaluate whether our process-oriented spatial reasoning correction complements inference-time spatial augmentation for final-answer performance, we combine it with SpatialPIN. As shown in Tab.\ref{tab:spatial_augmentation_combination}, the combination achieves an average accuracy of $65.45\%$, outperforming SpatialPIN alone ($56.95\%$) and our method alone ($64.75\%$). It also outperforms both methods across all three backbones. These results demonstrate the complementarity between inference-time spatial augmentation and process-oriented spatial reasoning correction.

\section{Related Work}

\subsection{Spatial Reasoning in Multimodal LLMs}


Existing methods for MLLM spatial reasoning mainly use training-based enhancement or inference-time augmentation. Training-based methods construct spatial instruction data or introduce region- and depth-aware representations~\cite{chen2024spatialvlm,cheng2024spatialrgpt,sun2025spatial,ogezi2025spare}. Inference-time methods provide depth cues, 3D priors, visual marks, or structured scene representations~\cite{ma2024spatialpin,roy2025bydeway,yang2023set,frisoni2026graph}. However, they enhance spatial knowledge without verifying whether intermediate judgments are visually supported. Our method instead extracts and verifies spatial evidence units from CoT and traces unsupported units to their source steps.

\begin{table}[t]
\centering
\small

\setlength{\tabcolsep}{4.8pt}
\renewcommand{\arraystretch}{1.10}

\resizebox{\columnwidth}{!}{%
\begin{tabular}{lcccc}
\toprule
\textbf{Method}
& \textbf{Qwen}
& \textbf{InternVL}
& \textbf{Llama}
& \textbf{Average} \\
\midrule

Vanilla
& 63.66
& 48.63
& 54.90
& 55.73 \\

SpatialPIN
& 64.44
& 60.52
& 45.88
& 56.95 \\

Ours
& 66.93
& 66.14
& 61.18
& 64.75 \\

\midrule


SpatialPIN + Ours
& \textbf{67.71}
& \textbf{66.41}
& \textbf{62.22}
& \textbf{65.45} \\

\bottomrule
\end{tabular}%
}

\captionsetup{skip=4pt}
\caption{
Performance of SpatialPIN, our process-oriented correction, and their combination on RealWorldQA.
}

\label{tab:spatial_augmentation_combination}
\vspace{-2ex}
\end{table}

\subsection{Faithful Reasoning and Verification}


Recent methods improve multimodal faithfulness through decoding correction, reasoning verification, or post-hoc grounding. Decoding methods suppress unsupported language priors by adjusting output distributions or strengthening visual grounding~\cite{wang2024mllm,ghosh2025visual}, while verifier-based and post-hoc methods check reasoning trajectories against localized visual evidence~\cite{sun2025mm,yi2025corgi}. FaithAct uses the visual faithfulness of intermediate reasoning to guide correction~\cite{li2026faithful}. However, these methods neither represent inter-object spatial relations as verifiable evidence units nor assess the reliability of corrective visual evidence. We address both limitations through SEG-based evidence tracing and SERA-based reliability-aware verification.

\section{Conclusion}
We propose a modular and training-free framework for perceptually faithful spatial reasoning in MLLMs. The SEG structures intermediate spatial evidence units and preserves their source steps, while SERA determines whether existence, localization, and geometric evidence is sufficiently reliable for verification. The framework then corrects the earliest contradicted evidence unit and regenerates the affected reasoning steps to prevent error propagation. 
Experiments across three representative MLLMs and five benchmarks show that our method achieves an average accuracy of 68.94\%, outperforming the compared baselines by 8.55 percentage points on average. These results demonstrate that evidence-level verification provides an effective approach to improving spatial reasoning.

\section*{Limitations}


Although in this work we present a systematic and training-free framework for verifying and correcting spatial reasoning in MLLMs, it still has several limitations. 

First, the framework relies on explicit CoT and can verify only the spatial evidence successfully extracted from the generated reasoning. Spatial errors that remain implicit or are omitted from the CoT cannot be directly localized or corrected, while non-spatial reasoning errors fall beyond the scope of the current framework. Second, our experiments mainly focus on static-image question-answering benchmarks and three open-source MLLM backbones. Its effectiveness in video understanding, multi-view reasoning, proprietary MLLMs, and closed-loop embodied interaction therefore remains to be investigated. Future work will explore the identification of implicit spatial evidence and extend the framework to a broader range of models, tasks, and dynamic embodied environments.



\bibliography{custom}

\clearpage

\appendix

\section{Discussion: Improving Accuracy through Perceptual Faithfulness}

Our analysis reveals a strong association between perceptual faithfulness and final-answer accuracy: reasoning chains containing contradicted spatial evidence are substantially less accurate than fully faithful ones. This observation motivates our central hypothesis that improving the perceptual faithfulness of intermediate spatial reasoning can lead to more accurate final answers. To operationalize this idea, SEG makes spatial evidence units explicit and traceable to their source steps, while SERA determines whether the corresponding visual evidence is sufficiently reliable for verification and correction. By locating the earliest contradicted evidence unit and regenerating the affected reasoning, our framework reduces the propagation of visually unsupported spatial judgments. The consistent improvements in both Spatial Evidence Faithfulness and final-answer accuracy across datasets and backbones provide empirical support for this hypothesis. Moreover, the gains obtained by combining our framework with SpatialPIN suggest that enriching spatial information and improving the faithfulness of generated reasoning are complementary: the former provides additional spatial cues, whereas the latter verifies whether the intermediate evidence actually used by the model is consistent with the image. These findings indicate that perceptual faithfulness is not only a property for evaluating spatial reasoning, but also a practical direction for improving its final-answer accuracy.

\section{Spatial Evidence Extraction and SEG Construction Details}
\label{app:evidence_extraction_prompt}

\subsection{Step-wise Spatial Evidence Extraction}

Given an initial reasoning chain
$C=\{c_1,c_2,\ldots,c_n\}$, we process each reasoning step $c_k$ to extract atomic spatial evidence units. Each extracted unit retains its source-step index $k$, allowing contradicted evidence to be traced back to its original position in the CoT. Following the definition in Sec.~3.1, the extractor covers four evidence types: object existence, absolute image position, 2D relative spatial relations, and relative depth.

Each spatial evidence unit is represented in the unified form
\[
s_m=\left\langle o_i,r_m,o_j,t_m,k_m\right\rangle,
\]
where $o_i$ and $o_j$ denote the involved visual entities or the image-reference node, $r_m$ denotes the encoded spatial relation, $t_m$ denotes the evidence type, and $k_m$ records the source reasoning step. For object-existence evidence, $r_m$ records whether the entity is stated to be present or absent and $o_j$ is not required. For absolute-position evidence, $o_j$ is the image-reference node. For 2D relative and relative-depth evidence, $o_i$ and $o_j$ denote the two involved visual entities.

The extractor records only spatial evidence explicitly expressed in the corresponding reasoning step. A mere mention of an entity is not treated as an object-existence judgment, and hypothetical, conditional, interrogative, or uncertain expressions are not converted into decisive existence evidence. Composite spatial statements are decomposed into atomic units whenever they contain multiple relations. For example, ``the lamp is to the right front of the vase'' is decomposed into a 2D relation (\texttt{right\_of}) and a relative-depth relation (\texttt{closer\_than}).

Before graph construction, relation labels are canonicalized to the predefined vocabulary, while entity expressions required to distinguish different visual instances are preserved. Spatial statements appearing in different reasoning steps remain separate evidence units because their source-step indices differ. Only exact duplicate outputs within the same reasoning step are removed.

\subsection{SEG Instantiation}

The extracted evidence units are organized into a Spatial Evidence Graph
$G=(V,E,A)$. The node set $V$ contains visual entity nodes and an additional image-reference node. The edge set $E$ contains directed spatial-relation edges. The attribute set $A$ stores the evidence type, source-step index, and visual evidence associated with each node or edge.

Object-existence evidence is stored as an attribute of the corresponding entity node. Absolute-position evidence is represented as a directed edge from an entity node to the image-reference node. Both 2D relative relations and relative-depth relations are represented as directed edges between visual entity nodes. Each evidence record preserves its relation type and source-step index, enabling chronological ordering and subsequent error localization.

\subsection{Visual Evidence Binding}

For each visual entity involved in the extracted evidence units, Grounding DINO provides candidate bounding boxes and their confidence scores. The highest-scoring candidate is used as the primary localization result, while the top two raw grounding scores are retained for the localization-reliability computation defined in Sec.~3.2. Given a selected bounding box
$b_o=(x_1,y_1,x_2,y_2)$, its center is
\[
x_c(o)=\frac{x_1+x_2}{2},
\qquad
y_c(o)=\frac{y_1+y_2}{2}.
\]

The image-level existence probability $p_{\mathrm{yes}}(o)$ is computed following FaithAct and is bound to the corresponding entity node for object-existence verification. Bounding-box coordinates and center locations provide the geometric evidence used for absolute-position and 2D relative-relation evidence units.

For relative-depth evidence units, the selected bounding box is used to prompt SAM2 to obtain an object mask. A monocular depth estimator then provides the depth values within the masked region, from which the object-level depth statistic and within-region stability information are computed. These measurements are bound to the corresponding depth-relation edge for subsequent reliability assessment and verification.

The resulting SEG therefore explicitly links each atomic spatial evidence unit to the visual observations required for verification. The reliability computation, verification decisions, and targeted correction report used by SERA are described separately in the reliability-aware verification appendix.

\subsection{Spatial Evidence Extraction Prompt}

Tab.~\ref{tab:evidence_extraction_prompt} presents the prompt used to extract step-level spatial evidence units. The extractor is applied to each reasoning step, and the resulting units are subsequently combined to instantiate the SEG.

\begin{table*}[t]
  \centering
  \small
  \begin{tabular}{p{0.95\textwidth}}
    \toprule
    \textbf{Prompt used for step-level spatial evidence extraction} \\
    \midrule

    Read the reasoning step below from a vision QA chain and extract all atomic spatial evidence units stated in this step. Do not use information from other reasoning steps and do not invent unstated relations. \newline

    Each evidence unit must contain: \newline
    \{subject, relation, object, type, source\_step\}. \newline

    Supported evidence types and relations: \newline
    1. \textbf{existence}: \texttt{present}, \texttt{absent}. Set \texttt{object} to null. \newline
    2. \textbf{absolute\_position}: \texttt{left}, \texttt{right}, \texttt{top}, \texttt{bottom}, \texttt{center}, \texttt{top\_left}, \texttt{top\_right}, \texttt{bottom\_left}, \texttt{bottom\_right}. Set \texttt{object} to \texttt{image}. \newline
    3. \textbf{relative\_2d}: \texttt{left\_of}, \texttt{right\_of}, \texttt{above}, \texttt{below}, \texttt{near}. \newline
    4. \textbf{relative\_depth}: \texttt{closer\_than}, \texttt{farther\_than}. Map ``in front of'' to \texttt{closer\_than} and ``behind'' to \texttt{farther\_than}. \newline

    Rules: \newline
    - Copy the integer from the \texttt{[Step N]} label into \texttt{source\_step}. \newline
    - Preserve the entity expression needed to identify the intended visual instance, such as ``the person on the left'' or ``the second car''. \newline
    - Extract an existence unit only when the step explicitly judges that an entity is present or absent. \newline
    - Do not treat a mere entity mention as an existence judgment. \newline
    - Do not convert hypothetical, conditional, interrogative, or uncertain expressions into decisive evidence units. \newline
    - Decompose a statement containing multiple spatial relations into separate atomic evidence units. \newline
    - Include repeated statements when they occur in different reasoning steps; their source steps are different. \newline
    - If the step contains no spatial evidence, return an empty array. \newline

    REASONING STEP: \newline
    \#\#\#\{CURRENT CoT STEP WITH [Step N] LABEL\}\#\#\# \newline

    Return ONLY valid JSON with no Markdown formatting: \newline
    \{
    ``evidence\_units'': [
    \{
    ``subject'': ``the bus'',
    ``relation'': ``left\_of'',
    ``object'': ``the car'',
    ``type'': ``relative\_2d'',
    ``source\_step'': 1
    \},
    \{
    ``subject'': ``the person on the left'',
    ``relation'': ``closer\_than'',
    ``object'': ``the person on the right'',
    ``type'': ``relative\_depth'',
    ``source\_step'': 1
    \}
    ]
    \} \\

    \bottomrule
  \end{tabular}
  \caption{Prompt used to extract atomic spatial evidence units from each reasoning step.}
  \label{tab:evidence_extraction_prompt}
\end{table*}

\section{Verification Report and Reasoning Regeneration Details}
\label{app:reasoning_regeneration_prompt}

\subsection{Spatial Evidence Verification Rules}

For every visual entity involved in the reasoning chain, we first compute the predefined image-level existence judgment $J_E(o)$. This node-level judgment is used both to verify explicit object-existence evidence units and to check the entity-existence presuppositions of absolute-position, 2D relative-relation, and relative-depth evidence units. The node-level judgment itself does not introduce an additional existence evidence unit unless the initial CoT explicitly expresses the corresponding existence judgment.

Explicit object-existence evidence units are verified according to their extracted polarity. A \texttt{present} unit is marked as \textsc{Supported} when $J_E(o)=\textsc{Present}$ and as \textsc{Contradicted} when $J_E(o)=\textsc{Absent}$. The decisions are reversed for an \texttt{absent} unit. If $J_E(o)=\textsc{Uncertain}$, the evidence unit is marked as \textsc{Inconclusive}. Verification of an existence unit depends only on the existence judgment determined by $R_E(o)$ and $p_{\mathrm{yes}}(o)$, rather than on the aggregated entity-reliability tier.

For absolute image position, the bounding-box center is first normalized by the image width $W$ and height $H$:
\[
\tilde{x}_c(o)=\frac{x_c(o)}{W},
\qquad
\tilde{y}_c(o)=\frac{y_c(o)}{H}.
\]

The normalized center is then mapped to fixed horizontal and vertical partitions corresponding to \texttt{left}, \texttt{center}, and \texttt{right}, together with \texttt{top}, \texttt{center}, and \texttt{bottom}. Their combinations produce the nine supported image regions. The same frame-partition rules are used across all datasets and models.

Pairwise left--right and above--below relations are determined using bounding-box centers. Since the image-coordinate origin is located at the upper-left corner, a smaller $x$ coordinate indicates a more leftward position, while a smaller $y$ coordinate indicates a higher position:
\[
o_i\ \texttt{left\_of}\ o_j
\iff
x_c(o_i)<x_c(o_j),
\]
\[
o_i\ \texttt{above}\ o_j
\iff
y_c(o_i)<y_c(o_j).
\]
The inverse comparisons are used for \texttt{right\_of} and \texttt{below}. A fixed spatial margin is applied in implementation to avoid decisive judgments when two centers are nearly aligned.

The \texttt{near} relation is evaluated using the normalized 2D distance between the two object centers:
\[
\begin{aligned}
d_{\mathrm{2D}}(o_i,o_j)
&=
\Biggl[
\left(
\frac{x_c(o_i)-x_c(o_j)}{W}
\right)^2 \\
&\quad+
\left(
\frac{y_c(o_i)-y_c(o_j)}{H}
\right)^2
\Biggr]^{1/2}.
\end{aligned}
\]
A pair is considered near only when this distance is below the fixed threshold used in our implementation.

For relative-depth relations, $z(o)$ denotes the mask-based object-level depth statistic computed after reversing the ordering of the raw inverse-depth prediction. Consequently, a smaller $z(o)$ indicates that an object is closer to the camera, whereas a larger value indicates that it is farther:
\[
o_i\ \texttt{closer\_than}\ o_j
\iff
z(o_i)<z(o_j),
\]
\[
o_i\ \texttt{farther\_than}\ o_j
\iff
z(o_i)>z(o_j).
\]
A fixed depth-difference margin is used to avoid decisive verification when the estimated depths are too similar. Monocular depth is compared only between objects in the same image.

For an absolute-position, 2D relative-relation, or relative-depth evidence unit, entity-existence presuppositions are checked before geometric verification. If any involved entity has $J_E(o)=\textsc{Absent}$, the evidence unit is marked as \textsc{Contradicted} because its existence presupposition is violated. If any involved entity has $J_E(o)=\textsc{Uncertain}$, the evidence unit is marked as \textsc{Inconclusive}. Geometric verification is performed only when all involved entities have $J_E(o)=\textsc{Present}$. In this case, the corresponding geometric rule produces a \textsc{Supported} or \textsc{Contradicted} result only when every involved entity is assigned to the HIGH tier. If any involved entity is assigned to the MEDIUM or LOW tier, the evidence unit is marked as \textsc{Inconclusive}.

\subsection{Verification Report Construction}

After evidence-level verification, we construct a verification record for each evidence unit containing its textual content, source reasoning step, verification result, evidence-specific reliability information, and precomputed visual judgment. All records are ordered by their source-step indices and, within the same step, by their order of occurrence in the initial CoT.

For each entity, the report retains its image-level existence score $p_{\mathrm{yes}}(o)$, existence reliability $R_E(o)$, and precomputed existence judgment. For an explicit object-existence evidence unit, the report additionally retains its extracted polarity and source reasoning step. The existence judgment is defined as
\[
J_E(o)=
\begin{cases}
\text{\textsc{Present}},
& \substack{
R_E(o)\geq\tau_h\\
p_{\mathrm{yes}}(o)>0.5
},\\[2mm]
\text{\textsc{Absent}},
& \substack{
R_E(o)\geq\tau_h\\
p_{\mathrm{yes}}(o)<0.5
},\\[2mm]
\text{\textsc{Uncertain}},
& R_E(o)<\tau_h.
\end{cases}
\]

Here, \textsc{Present} and \textsc{Absent} denote reliable presence and reliable absence, respectively, whereas \textsc{Uncertain} indicates that the available existence evidence is insufficient for a decisive judgment. This judgment is node-level visual evidence computed for each relevant entity and does not convert a mere entity mention into an explicit existence evidence unit. The regeneration model directly uses this precomputed judgment rather than inferring entity existence again from $p_{\mathrm{yes}}(o)$.

For 2D relative relations, the report provides a precomputed \textit{Layout judgment}; for absolute image positions, it provides a \textit{Frame judgment}; and for relative-depth relations, it provides a precomputed \textit{Depth judgment}. A geometric judgment is treated as decisive only when all involved entities are reliably determined to be present and are assigned to the HIGH tier. If an absolute-position or relation evidence unit is contradicted because an involved entity is reliably determined to be absent, the report explicitly identifies the violation of its entity-existence presupposition. If entity existence is uncertain or geometric reliability is insufficient, the evidence unit is reported as \textsc{Inconclusive}. The reasoning model directly uses these precomputed judgments rather than reinterpreting raw bounding boxes, grounding scores, or depth values.

Let the set of evidence units contradicted by reliable visual evidence be
\[
\mathcal{S}^{-}
=
\left\{
s_m
\mid
\operatorname{ver}(s_m)=\textsc{Contradicted}
\right\}.
\]

If this set is nonempty, we select the earliest contradicted evidence unit according to the provenance index $k_m$ stored in the SEG:
\[
s^{\star}
=
\underset{s_m\in\mathcal{S}^{-}}{\arg\min}\; k_m.
\]
If multiple contradicted evidence units share the earliest source-step index, the first unit according to their original order within that reasoning step is selected.

The earliest contradicted evidence unit $s^{\star}$ is highlighted in the verification report, while relevant \textsc{Supported} and \textsc{Inconclusive} judgments are also retained. \textsc{Supported} results help the model preserve already correct judgments, whereas \textsc{Inconclusive} results indicate that the available evidence is insufficient and should not be treated as mandatory correction signals.

\subsection{Targeted Reasoning Regeneration}

When a reliable contradiction is found, the original image, original question, initial reasoning chain, initial answer, verification report, and dataset-specific answer-format requirements are jointly provided to the same MLLM that generated the initial reasoning. The model is not required to reinterpret the low-level visual evidence; instead, it revises the localized error according to the verified spatial judgments in the report.

During regeneration, the model is instructed to preserve judgments consistent with reliable evidence, revise the earliest contradicted evidence unit $s^{\star}$, and regenerate the affected reasoning from its source step together with the final answer. If a spatial-relation evidence unit is contradicted because an involved entity is reliably determined to be absent, the model should remove or revise the unsupported relation rather than infer the opposite spatial relation.

The model is not required to repeat existence probabilities, reliability scores, bounding boxes, grounding scores, depth values, or other low-level evidence in its output. Evidence units marked as \textsc{Inconclusive} must not be treated as decisive correction signals.

When no reliable contradiction is identified, i.e., $\mathcal{S}^{-}=\varnothing$, regeneration is not performed, and the initial reasoning chain and answer are retained. This fallback strategy prevents unnecessary modifications caused by MEDIUM- or LOW-reliability evidence or uncertain existence evidence.

The entire correction process is performed using the original MLLM and requires no task-specific model fine-tuning or parameter updates.

\subsection{Reasoning Regeneration Prompt}

The complete prompt used for targeted reasoning regeneration is shown in Tab.~\ref{tab:reasoning_regeneration_prompt}. Its placeholders are replaced with the original question, initial reasoning, initial answer, generated verification report, and answer-format requirements of the corresponding dataset.

\begin{table*}[t]
\centering
\footnotesize
\begin{tabular}{p{0.95\textwidth}}
\toprule
\textbf{Prompt used for targeted reasoning regeneration} \\
\midrule

You are given an image, a question, your previous reasoning and final answer, and a verification report containing spatial judgments derived from reliable visual evidence. \newline

Question: \newline
\{Original question\} \newline

Previous reasoning: \newline
\{Previous reasoning\} \newline

Previous final answer: \newline
\{Previous final answer\} \newline

Verification Report: \newline
\{Verification Report, i.e., additional\_information\} \newline

Report interpretation rules: \newline
- Follow the precomputed Existence judgment for each entity: PRESENT indicates reliable presence, ABSENT indicates reliable absence, and UNCERTAIN indicates insufficient existence evidence. \newline
- Do not infer that every entity mentioned in the report is present in the image. \newline
- Use the precomputed Layout judgment to determine left/right/above/below relations. \newline
- Use the precomputed Frame judgment to determine an object's absolute position in the image. \newline
- Use the precomputed Depth judgment to determine closer/farther or front/behind relations. \newline
- Do not reinterpret or recompute a judgment from the reported probability, reliability score, bounding box, grounding score, or depth value. \newline
- A SUPPORTED result should be preserved unless it is affected by the correction of an earlier contradicted evidence unit. \newline
- A CONTRADICTED result should be revised according to the corresponding precomputed visual judgment. \newline
- If a relation is CONTRADICTED because an involved entity is ABSENT, remove or revise the unsupported relation; do not infer the opposite spatial relation. \newline
- An INCONCLUSIVE result is not a mandatory correction signal. Do not force a spatial judgment from inconclusive evidence. \newline
- Do not replace a 2D left/right relation with a depth or front/behind relation, or vice versa. \newline

Correction instructions: \newline
- Revise the highlighted earliest contradicted evidence unit and regenerate the affected reasoning from its source step together with the final answer. \newline
- Preserve earlier reasoning steps and other judgments that remain consistent with the verified evidence. \newline
- Ensure that the regenerated reasoning and final answer are mutually consistent. \newline
- Do not introduce unsupported objects, attributes, or spatial relations. \newline
- Do not repeat the verification evidence, probabilities, reliability scores, bounding boxes, grounding scores, or depth values in your reasoning. \newline
- Think step by step. \newline

Preserve the original step-based reasoning format whenever possible, and return the revised reasoning followed by the final answer. \newline

\{Dataset-specific answer-format requirements\} \\

\bottomrule
\end{tabular}
\caption{Prompt used to guide targeted reasoning regeneration using the verification report.}
\label{tab:reasoning_regeneration_prompt}
\end{table*}

\section{Answer Evaluation Prompts}
\label{app:answer_evaluation_prompts}

\paragraph{Evaluation protocol.}
For RealWorldQA, GQA, and POPE, we first apply deterministic
answer-extraction and normalization rules. GPT-4o is invoked as
a text-only semantic evaluator only when the rule-based procedure
is inconclusive. For all GPT-4o evaluations, we set the temperature
to 0 and the maximum number of output tokens to 8, and use the same
system prompt. For RealWorldQA, GQA, and POPE, only the last 4,500
characters of the model response are retained for evaluation.

\paragraph{GPT-4o-based binary semantic accuracy.}
For LLaVA-Bench-in-the-Wild and MMHal-Bench, we do not use
exact string matching. Instead, all valid samples are evaluated
using GPT-4o-based binary semantic judgments. The evaluator returns
\textsc{Yes} when the response correctly answers the question and
is semantically consistent with the reference answer, and
\textsc{No} otherwise. Synonyms,
paraphrases, and harmless formatting or detail differences are
accepted, whereas materially incorrect answers, contradictions,
failures to answer, or unsupported statements that change the
answer's meaning are judged incorrect.

A \textsc{Yes} judgment is recorded as
\texttt{correct=true}, whereas a \textsc{No} judgment is recorded
as \texttt{correct=false}. The reported accuracy is the proportion
of successfully scored samples judged as \textsc{Yes}:
\[
\mathrm{Acc}_{\mathrm{GPT\text{-}4o}}
=
\frac{N_{\mathrm{correct}}}
{N_{\mathrm{scored}}},
\]
where \(N_{\mathrm{correct}}\) is the number of samples judged as
\textsc{Yes}, and \(N_{\mathrm{scored}}\) is the number of samples
for which a valid binary judgment is successfully obtained.
Samples that cannot be scored successfully are excluded from the
denominator.


\begin{table*}[t]
  \centering
  \small
  \begin{tabular}{p{0.95\textwidth}}
    \toprule
    \textbf{Common system prompt used for answer evaluation} \\
    \midrule

    Reply with exactly YES or NO. No punctuation or explanation. \\

    \bottomrule
  \end{tabular}
  \caption{Common system prompt used for GPT-4o-based answer evaluation.}
  \label{tab:common_evaluation_system_prompt}
\end{table*}


\begin{table*}[t]
  \centering
  \small
  \begin{tabular}{p{0.95\textwidth}}
    \toprule
    \textbf{Prompt used for evaluating LLaVA-Bench-in-the-Wild and MMHal-Bench} \\
    \midrule

    You are a strict correctness grader for \{benchmark\_name\}. \newline
    Evaluate the model response as a binary correct/incorrect
    visual-question-answering response. \newline
    Return YES when it correctly answers the question and is
    semantically consistent with the reference answer. \newline
    Synonyms, paraphrases, and harmless formatting or detail
    differences are acceptable. \newline
    The reference answer is not necessarily the only valid wording. \newline
    Return NO for a materially wrong answer, contradiction, failure
    to answer, or unsupported statements that change the answer's meaning.
    A clear factual contradiction anywhere in the response makes it
    incorrect.

    Question: \newline
    \{question\}

    Reference answer: \newline
    \{gt\}

    Model response: \newline
    \{model\_answer\}

    Reply with exactly one token: YES if correct, otherwise NO. \\

    \bottomrule
  \end{tabular}
  \caption{
  Binary semantic-correctness prompt used for LLaVA-Bench-in-the-Wild
  and MMHal-Bench. The placeholder \{benchmark\_name\} is replaced
  with the corresponding benchmark name.
  }
  \label{tab:llava_mmhal_evaluation_prompt}
\end{table*}


\begin{table*}[t]
  \centering
  \small
  \begin{tabular}{p{0.95\textwidth}}
    \toprule
    \textbf{Prompt used for evaluating RealWorldQA and GQA} \\
    \midrule

    You are a strict grader for vision-language QA. \newline
    Decide whether the model's answer expresses the SAME final
    choice as the ground truth. \newline
    - For multiple-choice, same option letter (or same wording as
    the labeled choice) counts as match. \newline
    - For two-option word questions (e.g. two nouns), treat synonyms
    or clear paraphrases as match only if they pick the same entity. \newline
    - Ignore harmless formatting differences.

    Question (may be empty): \newline
    \{question\}

    Ground truth (GT): \newline
    \{gt\}

    Model response (may include long chain-of-thought; judge the
    intended final answer): \newline
    \{model\_answer\}

    Reply with exactly one token: YES if equivalent, otherwise NO. \\

    \bottomrule
  \end{tabular}
  \caption{
  Semantic-equivalence prompt used for RealWorldQA and GQA when
  deterministic answer extraction and normalization are inconclusive.
  }
  \label{tab:realworldqa_gqa_evaluation_prompt}
\end{table*}


\begin{table*}[t]
  \centering
  \small
  \begin{tabular}{p{0.95\textwidth}}
    \toprule
    \textbf{Prompt used for evaluating POPE} \\
    \midrule

    You are a strict grader for vision-language QA. \newline
    Decide whether the model's answer expresses the SAME final
    choice as the ground truth. \newline
    - For multiple-choice, same option letter (or same wording as
    the labeled choice) counts as match. \newline
    - For two-option word questions (e.g. two nouns), treat synonyms
    or clear paraphrases as match only if they pick the same entity. \newline
    - Ignore harmless formatting differences.

    This is a POPE-style binary task: ground truth is exactly YES
    or NO (object existence in the image). Decide whether the model's
    intended final judgment matches that YES/NO, even if the model
    did not print the word ``yes'' or ``no'' clearly.

    Question (may be empty): \newline
    \{question\}

    Ground truth (GT): \newline
    \{gt\}

    Model response (may include long chain-of-thought; judge the
    intended final answer): \newline
    \{model\_answer\}

    Reply with exactly one token: YES if equivalent, otherwise NO. \\

    \bottomrule
  \end{tabular}
  \caption{
  Binary object-existence evaluation prompt used for POPE when
  deterministic answer extraction is inconclusive.
  }
  \label{tab:pope_evaluation_prompt}
\end{table*}

\section{Self-Revision Prompt}
\label{app:self_revision_prompt}

Tab.~\ref{tab:self_revision_prompt} presents the complete prompt used for the generic Self-Revision baseline.

\begin{table*}[t]
  \centering
  \small
  \begin{tabular}{p{0.95\textwidth}}
    \toprule
    \textbf{Prompt used for Self-Revision} \\
    \midrule
    You are given an image, a question, and your previous reasoning and final answer.
    \newline\newline
    Please carefully reconsider the previous reasoning based on the image and the question. Check whether each step is consistent with the visual content and whether the final answer follows from the reasoning.
    \newline\newline
    Do not assume that the previous reasoning or answer is incorrect. Preserve correct judgments and revise only the parts that you independently determine should be changed.
    \newline\newline
    Question:
    \newline
    \{question\}
    \newline\newline
    Previous reasoning:
    \newline
    \{previous\_cot\}
    \newline\newline
    Previous final answer:
    \newline
    \{previous\_answer\}
    \newline\newline
    Please provide your revised reasoning and final answer in the following format:
    \newline\newline
    Reasoning: \textless your revised reasoning\textgreater
    \newline
    Final Answer: \textless your final answer\textgreater
    \\
    \bottomrule
  \end{tabular}
  \caption{Complete prompt used for the generic Self-Revision baseline.}
  \label{tab:self_revision_prompt}
\end{table*}

\end{document}